\documentclass[10pt,letterpaper]{article}
\usepackage[top=0.85in,left=2.75in,footskip=0.75in,marginparwidth=2in]{geometry}

\usepackage[utf8]{inputenc}

\usepackage{cite}

\usepackage{algorithm2e}

\usepackage{nameref,hyperref}

\usepackage[right]{lineno}

\usepackage{microtype}
\DisableLigatures[f]{encoding = *, family = * }

\raggedright
\usepackage{changepage}

\usepackage[aboveskip=1pt,labelfont=bf,labelsep=period,singlelinecheck=off]{caption}

\makeatletter
\renewcommand{\@biblabel}[1]{\quad#1.}
\makeatother

\usepackage{lastpage,fancyhdr,graphicx}
\usepackage{epstopdf}
\fancyheadoffset[L]{2.25in}
\fancyfootoffset[L]{2.25in}

\usepackage{color}

\definecolor{Gray}{gray}{.25}

\usepackage{graphicx}
\usepackage{multirow}
\usepackage{amsmath}
\usepackage{amssymb}
\usepackage{caption}

\usepackage{sidecap}

\usepackage{wrapfig}
\usepackage[pscoord]{eso-pic}
\usepackage[fulladjust]{marginnote}
\reversemarginpar

\usepackage{tikz}
\usepackage{textcomp}
\usepackage{hyperref}
\usepackage{lipsum}

\newcommand\copyrighttext{%
  \footnotesize \textcopyright 2012 IEEE. Personal use of this material is permitted.
  Permission from IEEE must be obtained for all other uses, in any current or future
  media, including reprinting/republishing this material for advertising or promotional
  purposes, creating new collective works, for resale or redistribution to servers or
  lists, or reuse of any copyrighted component of this work in other works.}
\newcommand\copyrightnotice{%
\begin{tikzpicture}[remember picture,overlay]
\node[anchor=south,yshift=10pt] at (current page.south) {\fbox{\parbox{\dimexpr\textwidth-\fboxsep-\fboxrule\relax}{\copyrighttext}}};
\end{tikzpicture}%
}

\begin{document}
\vspace*{0.35in}


\begin{flushleft}
{\Large
\textbf\newline{Multiple Regularizations Deep Learning for Paddy Growth Stages Classification from LANDSAT-8}
}
\newline
\\
Ines Heidiani Ikasari\textsuperscript{1}, 
Vina Ayumi\textsuperscript{1}, 
Mohamad Ivan Fanany\textsuperscript{1}, 
Sidik Mulyono\textsuperscript{2},
\\
\bigskip
\bf{1} Machine Learning and Computer Vision Laboratory \\Faculty of Computer Science, Universitas Indonesia\\
\bf{2} Center of Technology for Regional Resources Development (CTRRD) Agency for the Assessment and Application of Technology (BPPT)\\
\bigskip
* \{ines.heidiani,vina.ayumi\}@ui.ac.id

\end{flushleft}

\copyrightnotice

\providecommand{\keywords}[1]{\textbf{\textit{Keywords---}} #1}

\section*{Abstract}
This study uses remote sensing technology that can provide information about the condition of the earth's surface area, fast, and spatially.  The study area was in Karawang District, lying in the Northern part of West Java-Indonesia. We address a paddy growth stages classification using LANDSAT 8 image data obtained from multi-sensor remote sensing image taken in October 2015 to August 2016. This study pursues a fast and accurate classification of paddy growth stages by employing multiple regularizations learning on some deep learning methods such as DNN (Deep Neural Networks) and 1-D CNN (1-D Convolutional Neural Networks). The used regularizations are Fast Dropout, Dropout, and Batch Normalization. To evaluate the effectiveness, we also compared our method with other machine learning methods such as (Logistic Regression, SVM,  Random Forest, and XGBoost). The data used are seven bands of LANDSAT-8 spectral data samples that correspond to paddy growth stages data obtained from i-Sky (eye in the sky) Innovation system. The growth stages are determined based on paddy crop phenology profile from time series of LANDSAT-8 images. The classification results show that MLP using multiple regularization Dropout and Batch Normalization achieves the highest accuracy for this dataset. 
\bigskip

\noindent\keywords{Classification, Paddy Growth Stage, Remote Sensing, LANDSAT 8, Machine Learning, Deep Learning, Fast Dropout Training}


\section*{Introduction}
\label{Introduction}

Paddy is an important plant to Indonesian people as a staple food source. The needs of Indonesian national food is approximately 33-38 million tons of rice paddy per year. So, the government continues the effort to keep the stability of the fulfillment of basic needs of food, involving all relevant stakeholders, ranging from government policy, legal protection, supervision of agricultural production facilities in the market, technical assistance, to control the selling price in the market.

More reliable and rapid harvest yield estimation for paddy fields is also a critical issue to support the National Food Security Program that have been promoted and coordinated by the Indonesian government. 
Besides, an accurate and timely rice conditions, monitoring, and rapid rice harvest area estimation are certainly needed (\cite{Mulyono1}, \cite{Mulyono2}). Harvest yield estimation using conventional methods by directly measuring field are very subjective, costly, and spend a lot of time and effort \cite{Reynolds}. The main purpose of the observations using eye observation during the rice crop planting is to determine how much rice crop production in the future. However, since the limitations of human vision to observe in a relatively wide area, this way is not only subjective but also raises the data that tend to be over-estimated. On the other hand, rice planting schedules that are not uniform in every location also adds to the complex problems of rice in Indonesia. 

\par
An alternative solution for this issue is to use remote sensing technology, which is capable acquiring data in a vastly wide coverage area and in a short time. By using the approach of the rice plant phenology profile derived from remote sensing data series, it is proven that this technology is capable of monitoring paddy growth stages during its growing in each season with a good accuracy \cite{Mulyono3}. The detection of paddy growth stage is critical because much information can be extended from it, such as predictions of harvest time and the amount of harvest yield, the amount of water, seeds and fertilizer needed for the next planting season.

\section{Related Works}

Many studies on paddy phenology prediction using remote sensing data have been proposed by many researchers. Those studies use time-series of various vegetation indices, including Normalized Difference Vegetation Index (NDVI), Normalized Difference Soil Index (NDSI), Enhanced Vegetation Index (EVI), and Land Surface Water Index (LSWI) derived from MODIS images. The indices has effectively been applied for large scale mapping of flooding and transplanting paddy rice in summer season \cite{Xiao}.

Three procedures to determine paddy growth stages \cite{sakamoto}:
\begin{enumerate}
\item Prescription on constraining multi-temporal MODIS data
\item Filtering time-series EVI profile by time-frequency analysis
\item Specifying the phenological stages by detecting the maximum point, minimum point, and inflection point from the smoothed time-based EVI profile.
\end{enumerate}

Paddy growth stages are characterized by three stages from those procedures, (a) Heading date, (b) Flooding and planting date, (c) Harvesting date. For smoothing the EVI profile, we can use wavelet and Fourier transforms derived from MODIS satellite.

Another study \cite{Sari} proposed algorithm similar to \cite{sakamoto} for detecting paddy growth stage. The algorithm incorporates LSWI to determine the flooding date. Combination of LSWI and EVI was used to provide sensitivity to current water during flooding and transplanting date by using a global threshold value of 0.05 constrained by $LSWI+0.05 \geq EVI$ to detect flooding and transplanting date \cite{Xiao}.

Our previous studies used MODIS data to classify paddy growth stages. In this study, we use LANDSAT-8 data and compare some machine Learning methods, such as Logistic Regression, SVM,  Random Forest, XGBoost, MLP, and CNN for classifying paddy growth stages. And by adding some features like EVI, LSWI, NDVI, and ARVI (Atmospherically Resistant Vegetation Index), we trained and tested our data to obtain high accuracy.

\section{Methodology}
Regularization is an important technique in machine learning for dealing with overfitting problem without necessarilly adding more data. Basically, a regularization methods add one or more penalty terms in the objective function either analytically (e.g., Fast Dropout) or through indirect approach to randomized the learning system (e.g., Dropout) or the data (e.g., Batch Normalization). The regularization methods analyzed in this study are described in this following subsections.

\begin{figure}[h]
    \includegraphics[width=0.70\textwidth]{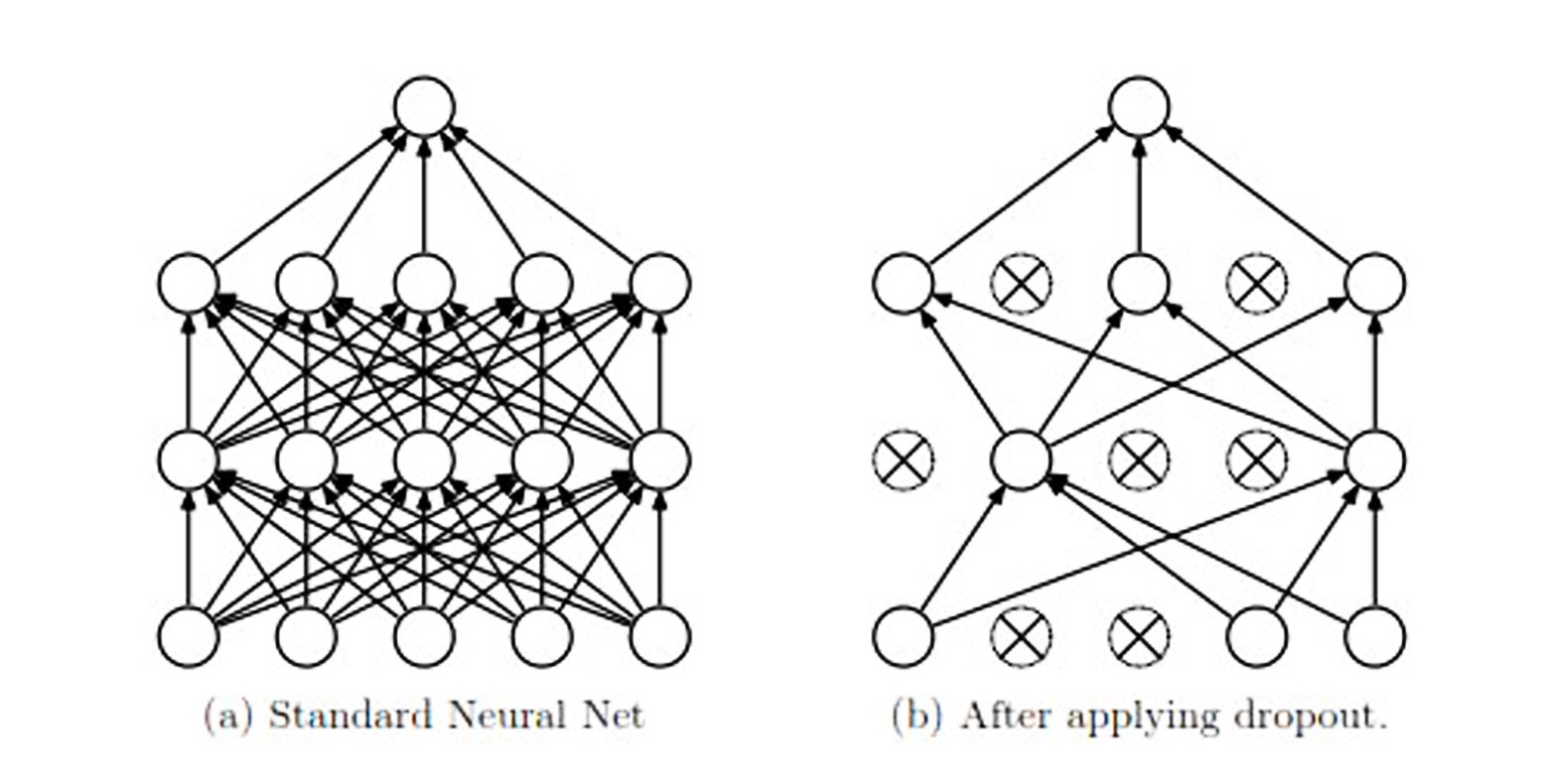}
    \caption{Dropout Neural Net Model \cite{Srivastava}}
    \label{fig:DO}
\end{figure}

\subsection{Dropout}
Hinton et al. introduced dropout training in the context of neural networks in \cite{Hinton}. It is a way to control overfitting in training procedure by randomly omitting subsets of features at each iteration. Dropout refers to dropping out units (hidden and visible) in a neural network. It prevents overfitting and provides a way of approximately combining exponentially many different neural network architectures efficiently. Dropping out or removing units from the network temporarily, along with all its incoming and outgoing connections is shown in Figure \ref{fig:DO}. The Figure shows a standard neural net with two hidden layers and neural networks before and after applying dropout. Applying dropout to a neural network amounts to sampling a “thinned” network from it. The thinned network consists of all the units that survived dropout. A neural net with $n$ units, can be seen as a collection of $2^n$ possible thinned neural networks. These networks all share weights so that the total number of parameters is still $O(n^2)$, or less. For each presentation of each training case, a new thinned network is sampled and trained. So training a neural network with dropout can be seen as training a collection of $2^n$ thinned networks with extensive weight sharing, where each thinned network gets trained very rarely, if at all \cite{Srivastava}. 

\subsection{Fast Dropout}

In the dropout training, for each data $\boldsymbol{\nu}$, we do Bernoulli sampling to generate a $k$ number of 0 or 1 random numbers. If the random number for a neuron is 1 then that neuron is activated (active), whereas if the random number is 0, then that neuron is not activated. The active neuron will be counted in calculating the output $h = w\cdot \boldsymbol{\nu} + b$. Depends on whether a particular neuron is active or not, it will be used in computing $h$ value during fast-forward phase and the $\partial h /\partial w$ in the back-forward phase. The behavior of Bernoulli random number for a large number of samples, according to Central Limit Theorem, approximately similar to the behavior of random number generated from Gauss or Normal distribution. This fact is 'exploited' by Fast Dropout techniques and has some interesting implications related to regularization. 

\begin{figure}[h]
    \includegraphics[width=0.7\textwidth]{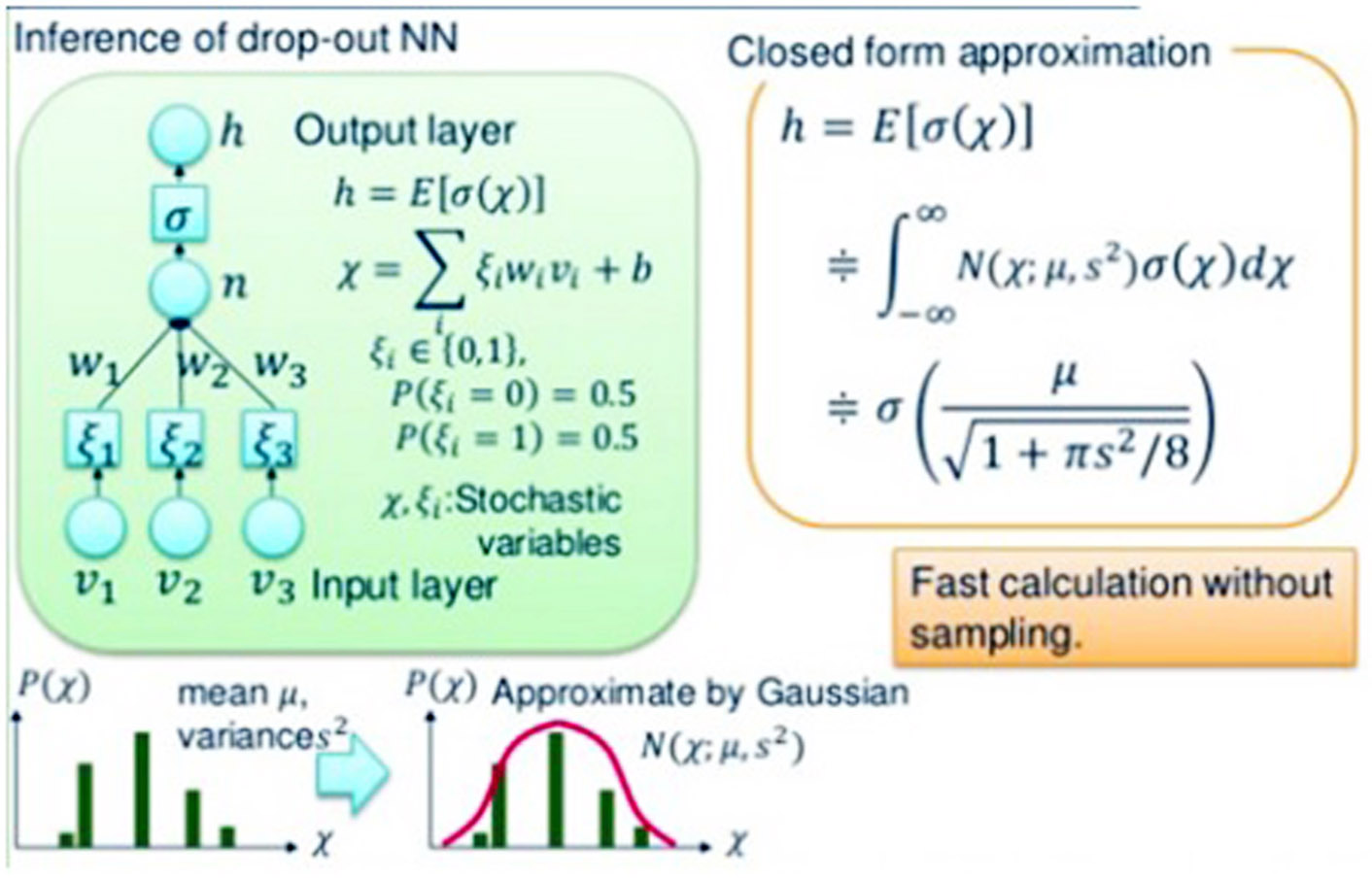}
    \caption{Approximation of Fast Dropout \cite{Tanaka}}
    \label{fig:fdo1}
\end{figure}

In fast dropout training, we do not perform Bernoulli sampling for each data $\boldsymbol{\nu}$, but we just collected the output $h$ from some data and computing its mean and variance. During the backward propagation phase to update the weights, we only randomly select the neuron from the $h$ distribution. The advantages of this fast dropout are in two fold. First, we only sample once before the weight updating process, not performing sampling every time the data $\boldsymbol{\nu}$ comes. Second, many data are "overlooked" due to dropout, but in fast dropout we just pass the data first and sample it next. It was shown that the fast dropout sometimes gives better results than the original dropout. Figure \ref{fig:fdo1} showed approximation of fast dropout \cite{Wang}.  

\begin{figure}[h]
    \includegraphics[width=0.5\textwidth]{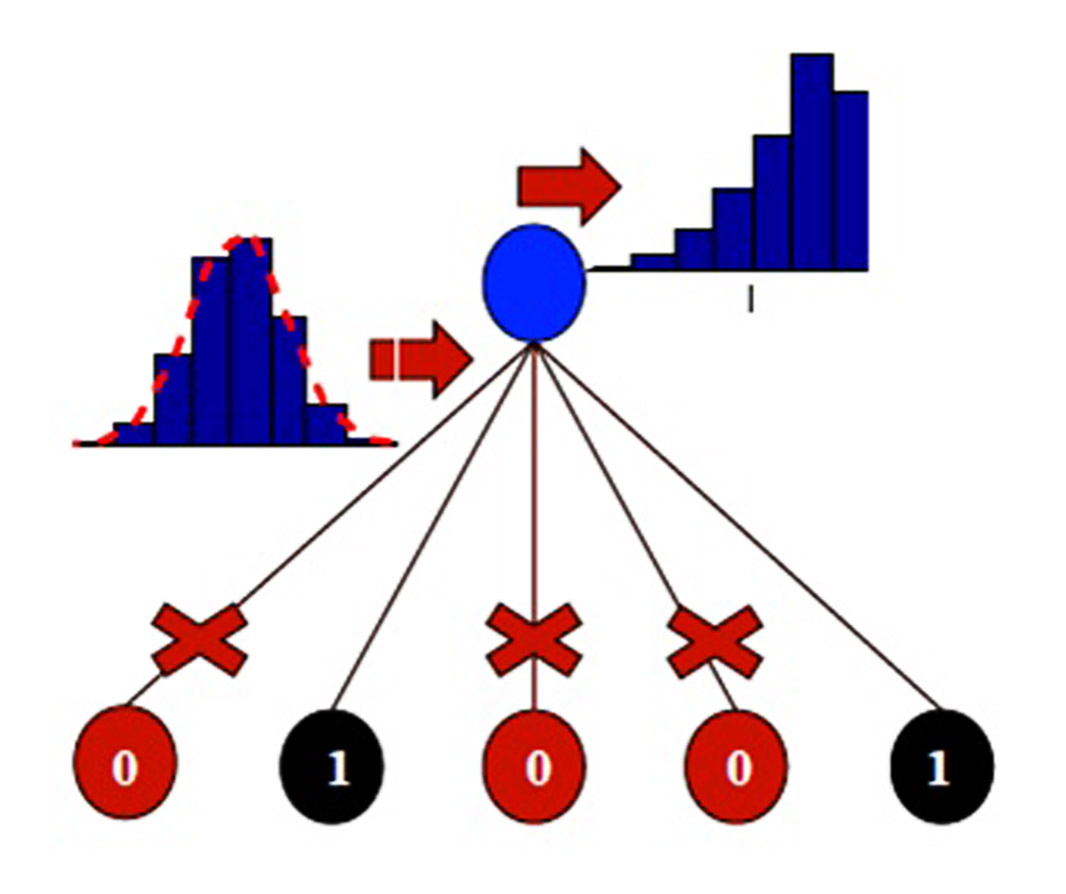}
    \caption{Illustration of Fast Dropout Idea \cite{Wang}}
    \label{fig:fdo2}
\end{figure}

Figure \ref{fig:fdo2} illustrated how fast dropout works. The numbers at the bottom are the dropout indicator variables $\nu_1$, ..., $\nu_5$.
As $\boldsymbol{\nu}$ is repeatedly sampled, the resulting inputs to the top unit are close to being normally distributed \cite{Wang}.

\subsection{Batch Normalization}

During training a deep neural network, the distribution of each layer's inputs changes as the parameters of the previous layers change. The layer's input distribution change requires lower learning rates and special parameter initialization, thus slows down the training. The layer's input distribution change is known as internal covariate shift. We can solve this internal covariate shift problem by normalizing layer inputs. The normalization can be performed for each training mini-batch and becomes a part of the model architecture. This normalization method is named as Batch Normalization (BN). 
The advantages of BN are allowing much higher learning rates and less susceptible to initialization. Besides, the BN also acts as a regularizer where in some cases eliminates the need for Dropout. \cite{Sergey} 

Fixing the distribution of the layer inputs as the training progresses might improve the training speed. It has long been known that the network training converges faster if its inputs are whitened, i.e., linearly transformed to have zero means and unit variances, and decorrelated. As each layer observes the inputs produced by the layers below, it would be desirable to have the same whitening of the inputs of each layer. Hence, each layer of the network could be whitened.  However, this often turns out to be computationally expensive.  

To approximates the whitening, batch normalization \cite{Sergey} uses the statistics of the current mini-batch by standardizing the intermediate representations. Given a mini-batch $x$, we calculate the sample mean and variance of each feature \( f \)  along the mini-batch axis, and \( n \) is the size of the mini-batch.

\begin{equation}
 \label{eq2}
     	\bar{x_f} = \frac{1}{n}  \sum_{i=1}^{n} x_{i,f},
\end{equation}

\begin{equation}
 \label{eq2}
     	\sigma^2_f = \frac{1}{n} \sum_{i=1}^{n} ((x_{i,f}-\bar{x_f})^2,
\end{equation}

\noindent Using these statistics, the each feature can be standardized as follows

\begin{equation}
 \label{eq3}
    \hat{x_f} = \frac{x_f - \bar{x_f}}{\sqrt{\sigma^2_f+\epsilon}}
\end{equation}

\noindent where \(\epsilon\) is a small positive constant to improve numerical stability. 

However, standardizing intermediate  activations can reduce representational power of the layer.  To address for this, BN introduces additional learnable parameters \(\gamma \) and \(\beta\), which respectively shift and scale the data, leading to a layer of the form.

\begin{equation}
 \label{eq4}
    BN(x_f) = \gamma_f \hat{x_f} + \beta_f
\end{equation}

The network can recover the original layer representation by setting \(\gamma_f\) to \(\sigma_f\) and \(\beta_f\) to \(\bar{x_f}\). Then for a standard feedforward layer in a neural network

\begin{equation}
 \label{eq5}
    y = \phi(W_x+b)
\end{equation}

\noindent where \(W\) is matrix of weights, \(b\) is the vector of bias, \(x\) is the input  of  the  layer  and \(\phi\) is a function of arbitrary  activation, the BN is applied as follows

\begin{equation}
 \label{eq6}
   y = \phi(BN(W_x))
\end{equation}

Note that the vector of bias has been removed, since its effect is canceled by the standardization.  After the normalization has been the part of the network, the procedure of back propagation needs to be adapted to propagate gradients through the mean and variance computations as well \cite{laurent}.

\section{Data and Experiments}
\subsection{Data Generation}

A series of LANDSAT-8 images used in this study are taken for free of charge from USGS Global Visualization Viewer page. A paddy growth stages map issued by i-Sky (eye in the sky) Innovation system belongs to BPPT was used as references to collect the data samples for training and testing. As an eye in the sky, satellite remote sensing in space provides information about the intensity of the reflection of sunlight on earth shown in the reflectance values of light with a wavelength range from the visible light to near infrared light. The reflectance values will be different for every object on earth, so this information can be used to recognize objects simultaneously monitor the growth of paddy rice. By using EVI (Enhanced Vegetation Index) and LSWI (land surface water index) within a particular time, it will obtain the curve that shows the pattern of the rice crop phenology profile \cite{Mulyono3}. These figures showed you Paddy phenological profile using EVI and LSWI.

\begin{figure}[h]
    \includegraphics[width=0.7\textwidth]{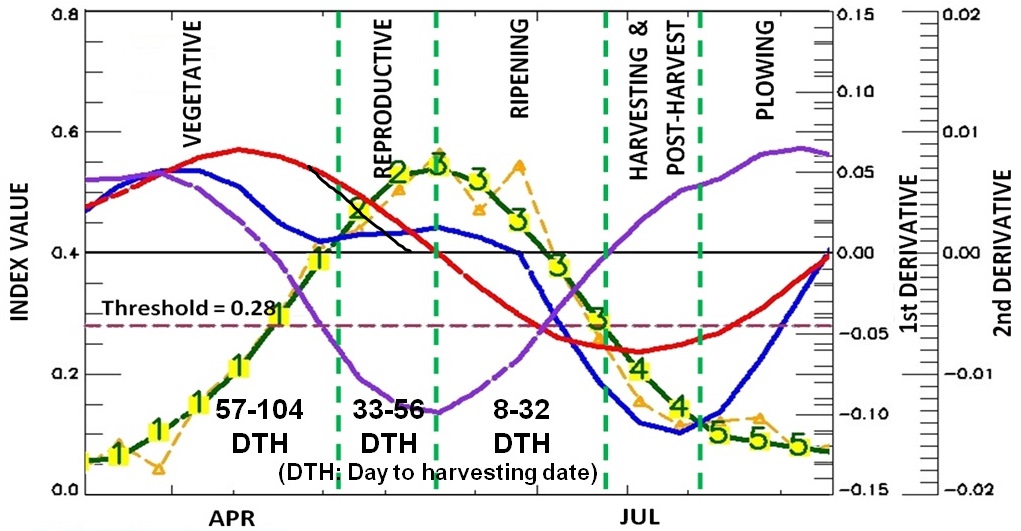}
    \caption{Visualization of Paddy Phenological Profile (Green line-EVI, Blue line-LSWI, Red line-First derivative of EVI, Purple line-Second derivative of EVI) \cite{Mulyono3}}
    \label{fig:visualizationofppp}
\end{figure}

\begin{figure}[h]
    \includegraphics[width=0.7\textwidth]{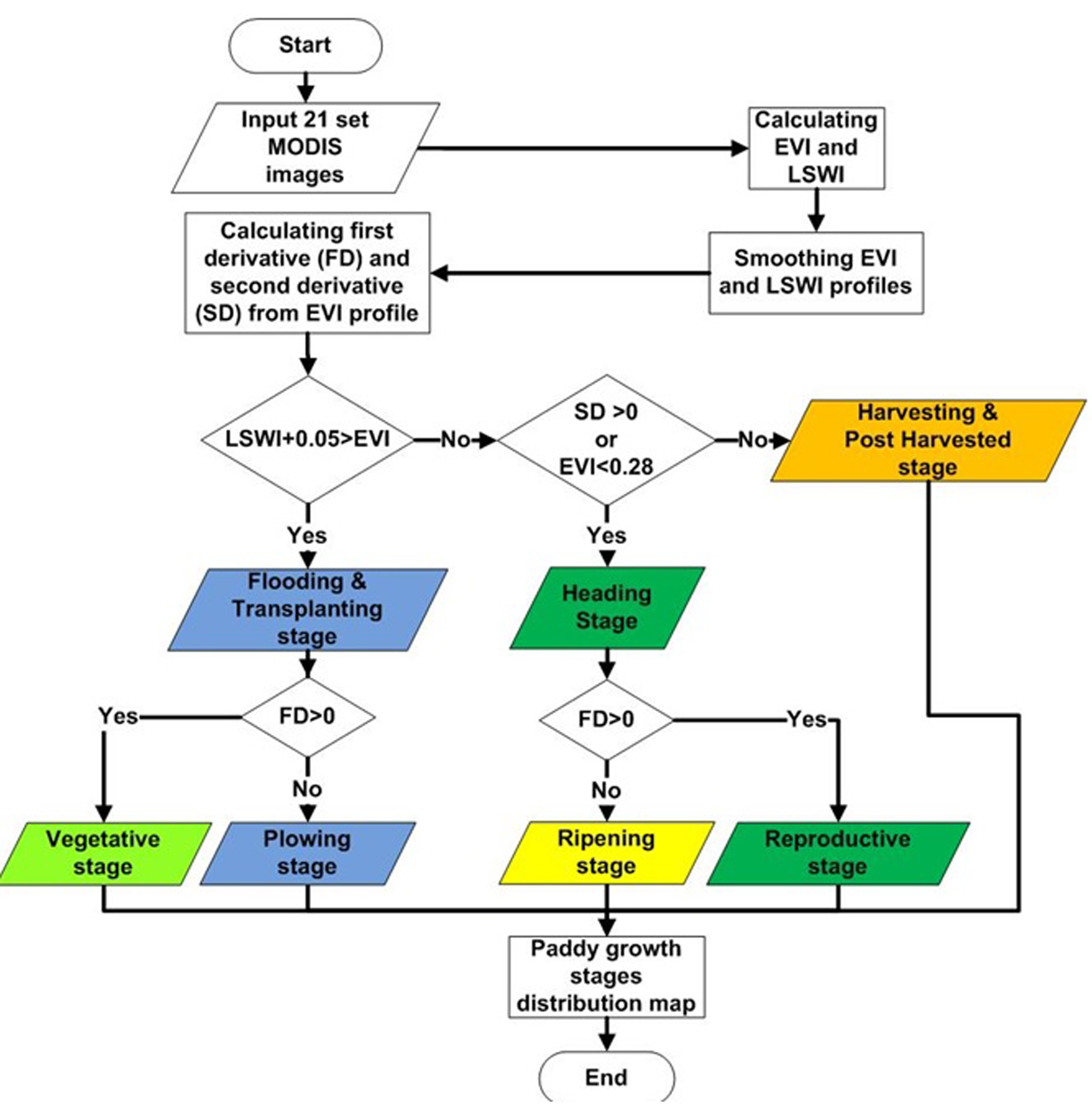}
    \caption{Heuristic Algorithm for Paddy Growth Stages Detection \cite{Mulyono3}}
    \label{fig:phenoflow}
\end{figure}

We used the terms of growth stages for paddy defined by i-Sky Innovation, i.e. vegetative, reproductive, ripening, harvesting and post harvesting, and plowing, which can be named to GS1, GS2, GS3, GS4, and GS5 respectively. Usually, the harvest period is quite short, so that these events are rarely be recorded by the image of LANDSAT-8 with 16 daily composites. Therefore, in this study, this stage is combined with the post-harvested stage. By using this way, we collected five classes for paddy growth stages from LANDSAT 8 images along paddy field area in Karawang District, lying in the Northern part of West Java-Indonesia. These pictures below were the result of paddy growth stages classification from i-Sky Innovation System at the meant time.

\begin{figure}[h]
    \includegraphics[width=1.0\textwidth]{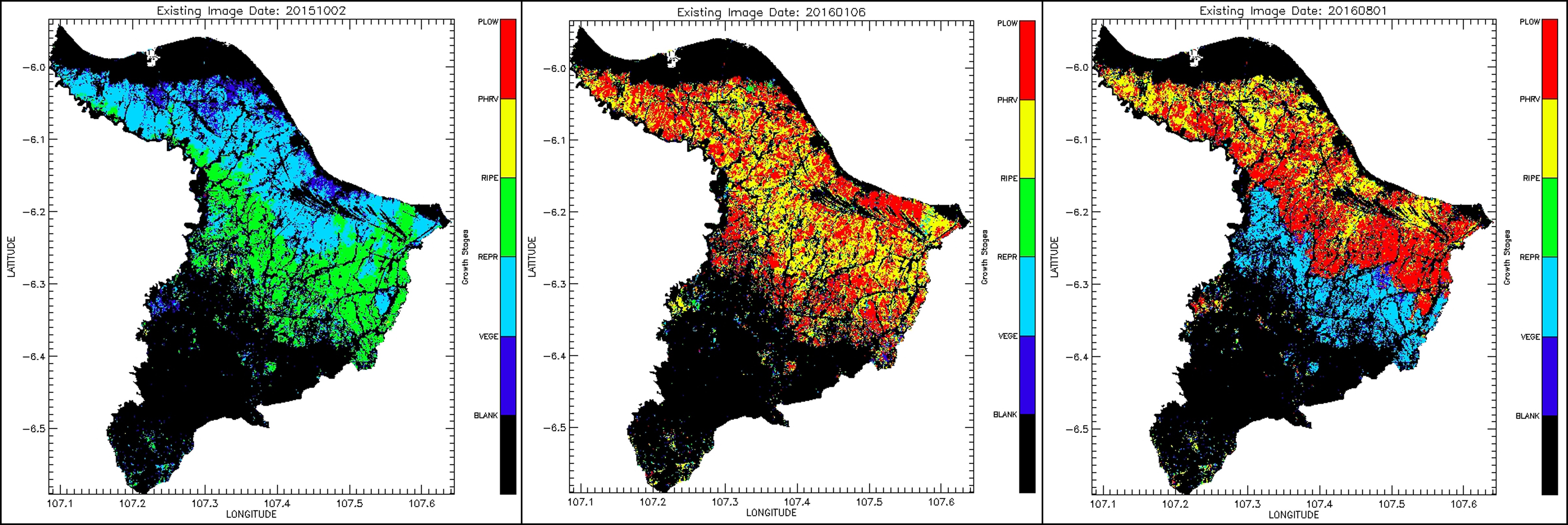}
    \caption{Paddy Growth Stage Classification from i-Sky Innovation System, Existing Image Date : (left) October 2nd, 2015, (middle) January 6th, 2016, (right) August 1st, 2016}
    \label{fig:137}
\end{figure}

\par
The total number of available data was collected from October 2nd, 2015 to August 1st, 2016. Before we could process these data, we should do preprocessing data. Preprocess data were done by using ENVI tools, such as:
\begin{enumerate}
    \item Radiometric Calibration 
    \item Dark Substraction 
    \item Atmospheric Correction 
    \item Vector to Region of Interest (ROI), in this study the ROI was Karawang, West Java, Indonesia 
    \item Convert to Geographic Lat/Lon 
\end{enumerate}

After we had done this preprocessing steps to our data, we chose the image that has no outliers and representing every stage from GS1 to GS5 to be used in next process. The selected images were on October 2nd, 2015, January 6th, 2016, and August 1st, 2016. Next process was separated fields with and without growth stages area. The fields with the growth stages area were known with the help of references of paddy field area from i-Sky Innovation system. 
It divided into five classes growth stages by using ArcMap tools. Moreover, by using ENVI Classic Tools, we collected data samples in text file format to be utilized in the process data with Deep Learning. These data samples still include area covered by cloud. Therefore we should remove it first. Table \ref{tab:one} showed the summary of the data samples after cloud removal.

\begin{table}[h]
    \caption{LANDSAT 8 Data Samples} 
    \bigskip
    \label{tab:one} 
    \begin{tabular}{|l|l|l|l|}
        \hline
No. & Date Taken & Growth Stage & Number of Samples \\ \hline
1   & 2015-10-02 & GS1 & 205,690 \\ 
2   & ~ & GS2 & 847,218 \\ 
3   & ~ & GS3 & 531,498 \\ 
4   & ~ & GS4 & 37,410 \\ 
5   & ~ & GS5 & 5,568 \\ \hline 6   & 2016-01-06 & GS1 & 78,411 \\ 
7   & ~ & GS2 & 80,940 \\ 
8   & ~ & GS3 & 55,582 \\ 
9   & ~ & GS4 & 1,179,142 \\ 
10  & ~ & GS5 & 975,878 \\ \hline
11  & 2016-08-01 & GS1 & 112,258 \\ 
12  & ~ & GS2 & 496,695 \\ 
13  & ~ & GS3 & 10,124 \\ 
14  & ~ & GS4 & 870,355 \\ 
15  & ~ & GS5 & 787,558 \\
        \hline
    \end{tabular}
\end{table}
These data samples have a different number of samples each stage in every date taken. To run with Deep Learning, we should balance it. Therefore, from all of the series data taken, we summed the number of data samples that have the same stage. Then, the stage which has less number of data samples than the other stages was used to balancing the number of data samples. Table \ref{tab:two} showed the change of the number of data samples.
\begin{table}[h]
    \caption{Balanced LANDSAT 8 Data Samples for Training and Testing with Deep Learning} 
    \bigskip
    \label{tab:two} 
    \begin{tabular}{|l|l|l|}
        \hline
        No. & Growth Stage  & Number of Samples \\ \hline
        1   & GS1           & 59,720            \\ 
        2   & GS2           & 59,720            \\ 
        3   & GS3           & 59,720            \\ 
        4   & GS4           & 59,720            \\ 
        5   & GS5           & 59,720            \\
        \hline
    \end{tabular}
\end{table}

After we had done the preprocessing steps, cloud removal, and balanced the data, next we do the feature extraction steps. We used 7 spectral band features and derived vegetation indices features from spectral band  combinations that are more representative of vegetation greenness. The indices include the Enhanced Vegetation Index (EVI \cite{EVI}), Normalized Difference Vegetation Index (NDVI \cite{NDVI}), Atmospherically Resistant Vegetation Index (ARVI \cite{ARVI}), and Land Surface Water Index (LSWI \cite{LSWI}).

\begin{equation}
 \label{eq2}
     	EVI = G \times \frac{NIR - Red}{NIR + c_{red}) \times Red - c_{blue}) \times Blue + L  }
\end{equation}
The coefficients \(G\),\(c_{red}\), \(c_{blue}\) and \(L\) are chosen to be 2.5, 6, 7.5 and 1 following those adopted in the MODIS EVI algorithm \cite{EVI}.

\begin{equation}
 \label{eq2}
     	NDVI = \frac{NIR-Red}{NIR+Red}
\end{equation}

\begin{equation}
 \label{eq2}
     	ARVI = \frac{NIR - (2 \times Red - Blue)}{NIR + (2 \times Red + Blue)}
\end{equation}

\begin{equation}
 \label{eq2}
     	LSWI = \frac{NIR-SWIR}{NIR+SWIR}
\end{equation}

\subsection{Experiments}

\subsubsection{Experimental Setup}
The experiment to classify paddy growth stage was conducted in Ubuntu 14.04 LTS 64, on a PC with  Processor Intel Core i7-5820K CPU 3.3 @Ghz, Memory DDR2 RAM 64.00 GB, Hard Disk 240 GB. In this study, we implemented some experiments using Deep Learning, i.e., Deep Neural Networks (DNN) and 1-D Convolutional Neural Networks (1-D CNN) with multiple regularizations, i.e., Dropout and Batch Normalization in Python using Keras deep learning library \cite{chollet2015keras}. For the DNN, we tried from four to six layers. Since the results of four layers are better than those of six layers, we only report the four layers DNN.   

As a comparison to our proposed method, we implemented other classifiers such as Logistic Regression (LR), SVM, Random Forest, XGBoost \cite{xgboost} (coded in Python) and Logistic Regression with Fast Dropout Training using fast dropout libray in MATLAB R2014a run on the same machine \cite{Wang}.

\subsubsection{Experimental Result}
By performing cross-validation, these number of data samples divided into data training and data testing. The training data took 2/3 of the dataset of each class and the rest of a testing data. Then, we could process these ready data samples to be trained and tested with deep and shallow learning methods. In the experiment using Deep Learning we used two layers (input and output) on DNN and CNN and added Batch Normalization on the data input layer. The Batch Normalization added before the activation, and the Dropout inserted after the activation. 

\begin{table}[h]
    \caption{Classification Accuracy} 
    \bigskip
    \label{tab:three}
    \begin{tabular}{|l|l|}
        \hline
        Machine Learning Methods & Accuracy (\%)\\ \hline
        LR + Fast Dropout        & 65.57        \\     
        SVM    & 60.65        \\ 
        Random Forest    & 65.34        \\ 
        XGBoost    & 68.36        \\ 
        DNN    & 69.29        \\ 
        DNN+Dropout    & 67.93        \\ 
        DNN+BN    & 68.57        \\ 
        DNN+BN+Dropout    & 71.79        \\
        CNN                      & 71.74        \\
        CNN+Dropout    & 70.54        \\ 
        CNN+BN    & 70.28        \\ 
        CNN+BN+Dropout    & 69.38        \\
        \hline
    \end{tabular}
    \label{tab:class_acc}
\end{table}

The experimental result showed in the Tabel \ref{tab:class_acc} below. The result indicates that DNN with BN+Dropout achieves higher accuracy than using single regularization (either Dropout or BN). On the other hand, CNN-only (without any regularizations) achieves higher accuracy than using any regurlarizations either BN, Dropout, BN+Dropout, and combined multiple regularizations. The comparison with another classifier shows that DNN with BN and Dropout has the highest accuracy than the other classifiers. Due to 1-D representation and weight sharing in CNN, the number of parameters (flexibility) seems lower than dense DNN. Given the same amount of data, the CNN seems can not adapt to data even when the regularizations are added. On the other hand, the DNN with more learned parameters (weights) still has room for adaptation and further regularizations seems work to bring the DNN from overfitting the training data. 

\par
After experiments, we could get a paddy growth stage classification by looking from these confusion matrices in Table 3 that contain information about actual and predicted classifications done by a classification system.

\section{Conclusion and Discussion}

This paper implements paddy growth stage classification using multiple regularizations learning which was carried out on two deep learning algorithms, i.e., CNN and DNN. This study uses the LANDSAT-8 image data obtained from the multisensor remote sensing image in Karawang District area. The experimental result shows that DNN with multiple regularizations, i.e., Dropout and Batch Normalization, gained higher accuracy compared with CNN and other traditional (shallow) machine learning classifiers. For future study, we suggest to explore further the fast dropout training in deep neural networks and use larger data samples. Time series predictions, which exploit sequential relations among serial remote sensing data, might also be an interest for further studies.


\bibliography{library}

\bibliographystyle{abbrv}

\end{document}